\documentclass[11pt]{article}

\usepackage[final]{acl}

\usepackage{times}
\usepackage{latexsym}
\usepackage{placeins}
\usepackage[flushmargin]{footmisc}
\usepackage[T1]{fontenc}
\usepackage[utf8]{inputenc}

\usepackage{microtype}

\usepackage{inconsolata}

\usepackage{graphicx}

\usepackage{subcaption}
\usepackage{array}
\usepackage{bbold}
\usepackage{multirow}
\usepackage{caption}
\usepackage{algorithm}      
\usepackage{algorithmic}
\usepackage{enumitem}
\usepackage{booktabs}
\usepackage{xcolor}
\usepackage{siunitx}
\usepackage{makecell}
\usepackage{amsmath}
\usepackage{tabularx}
\usepackage{colortbl}
\usepackage{upquote} 

\newcolumntype{Y}{>{\hsize=1.0\hsize\linewidth=\hsize}X}
\newcolumntype{Z}{>{\hsize=1.0\hsize\linewidth=\hsize}X}
\newif\ifcomment
\commentfalse

\title{Groundhog Bit-Flip Attack: Seeding Infinite Generation Loops in Mixture-of-Experts LLMs through Bit Flips}

\author{%
  Huakang Lin$^{1}$\thanks{\;Equal contribution.},
  Tiancheng Zheng$^{2*}$,
  Mingxuan Sun$^{3*}$,
  Tianhong Xu$^{4}$,
  Fan Zhang$^{5}$,
  Yunsi Fei$^{4}$,
  Ruyi Ding$^{1}$ \\[2pt]
  $^{1}$Louisiana State University, USA \quad
  $^{2}$University of California, Los Angeles, USA \\
  $^{3}$Southeast University, China \quad
  $^{4}$Northeastern University, USA \quad
  $^{5}$Zhejiang University, China \\[2pt]
  \texttt{hlin29@lsu.edu}, \texttt{zhengtiancheng@g.ucla.edu}, 
  \texttt{msun@seu.edu.cn},
  \texttt{ruyiding@lsu.edu}}

\begin{document}
\maketitle


\begin{abstract}

Mixture-of-Experts (MoE) architectures enable scalable and efficient large language models (LLMs) by selectively activating expert sub-networks through a routing mechanism.
However, this adaptive design introduces a new attack surface: specific experts become disproportionately correlated with certain tokens (e.g., end-of-sequence), allowing adversaries to manipulate model behavior via lightweight perturbations.
In this work, we present \textbf{Groundhog Bit-Flip Attack (GBFA)}, the first bit-flip-based \textit{ Denial-of-Wallet availability  attack} against MoE-based LLMs.
By identifying and flipping routing-layer bits associated with related expert activations, we demonstrate that GBFA substantially extends the decoding token usage across three different LLM modes: conversational, reasoning, and agentic tasks,  while largely preserving semantic fidelity.
Across four main real-world MoE-based LLMs, manually deactivating on average fewer than \textbf{4 experts} drives average output inflation to $\mathbf{5912\%}$, with the majority of test samples reaching max tokens.
These results reveal a robustness vulnerability of MoE architectures to bit flip, and highlight the potential of GBFA as an availability attack against LLMs.

\end{abstract}

\section{Introduction} \label{sec: introduction}


To reduce the inference cost of Large Language Models (LLMs), service providers have increasingly adopted Mixture-of-Experts (MoE) architectures, which expand model capacity while activating only a small subset of specialized experts for each token.
Recent models such as Mixtral~\cite{jiang2024mixtralexperts}, DeepSeek~\cite{deepseek_v2}, Qwen3-MoE~\cite{qwen3technicalreport}, and GPT-OSS~\cite{openai2025gptoss120bgptoss20bmodel} demonstrate the effectiveness of this design in supporting concurrent inference.
However, such sparse architecture also introduces new attack surfaces.

\begin{figure}[t]
    \centering
    \includegraphics[width=\linewidth]{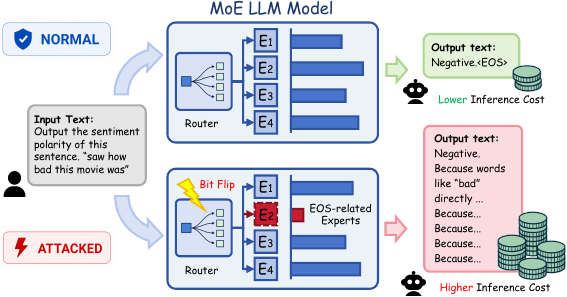}
    \caption{Groundhog Bit Flip Attack on MoE LLMs.}
    \label{fig:intro}
\end{figure}

 Prior works have focused on exploiting MoE-specific architectural vulnerabilities to attack LLM models. MoEcho~\cite{ding2025moecho} exploits temporal and spatial traces of expert execution as side channels to infer user prompts and outputs. SAFEx~\cite{lai2026safex} reveals that the intrinsic safety-aligned behaviors of MoE-based LLMs strongly depend on specific positional experts. In this work, we observe that the generation of termination tokens (e.g., end of sequence) is governed by a dedicated subset of experts. 
As a result, the same routing mechanism designed to improve inference efficiency can become an availability vulnerability: by disrupting these termination-related experts, an attacker can delay or suppress termination token generation. 
This threat is especially severe in rapidly growing agentic applications, where inflated outputs directly translate into \textbf{Denial-of-Wallet (DoW)} risk.

MoE sparsity makes bit-flip attacks a natural tool for availability attacks, as termination behavior can be concentrated in a small set of routing parameters. 
A few targeted bit flips can disrupt termination-related expert activations and induce excessive generation. 
However, existing BFA studies mainly focus on model integrity~\cite{rakin2022tbfa}, such as accuracy degradation or output corruption, while DoW attacks against MoE-based LLMs remain unexplored. 
To fill these gaps, we introduce \textbf{Groundhog Bit-Flip Attack (GBFA)}\footnote{Named after the film Groundhog Day. Throughout the paper, ``infinite'' denotes generation that exhausts the allocated token budget (\texttt{max\_new\_tokens}) rather than literally unbounded generation.}, the first bit-flip DoW attack on MoE-based LLMs.
GBFA treats MoE routing as a structural attack surface: as illustrated in Figure~\ref{fig:intro}, targeted bit flips in router parameters can reroute tokens away from termination-related experts, thereby delaying sequence termination.
This causes substantial output inflation and increased inference cost while largely preserving semantic fidelity.




Our analysis shows that MoE experts exhibit strong specialization for termination-related tokens, including end-of-sequence (EOS) and end-of-thinking (EOT) tokens. 
A small set of \emph{termination-related experts} disproportionately governs the generation of stops, making MoE routing a localized attack surface for availability manipulation. 
By applying router-level bit flips to disrupt these experts, an attacker can delay termination and induce excessive generation. 
This reframes bit-flip attacks from integrity-oriented attacks that degrade model accuracy or corrupt outputs into availability-oriented attacks that maximize output length while preserving apparent utility, directly leading to increased inference cost and Denial-of-Wallet risk.

This paper makes the following contributions:
\begin{itemize}[leftmargin=*]

\item \textbf{Expert–Token Specialization.} We discover specialization patterns in MoE experts, showing that a small set of experts dominate EOS or EOT handling and therefore govern termination behavior.

\item \textbf{Systematic Detection Methodology.} We introduce \textit{Global} and \textit{Local Expert Detection} that use activation-difference metrics to efficiently identify \textit{target-related experts} that are consistently activated when specific tokens generating.

\item \textbf{Identify Vulnerable Bits.} We show that a small number of router-weight or bias bits exert disproportionate control over target-related experts. By analyzing expert-activation sensitivity under bit-level perturbations, we identify Vulnerable Bits for GBFA.


\item \textbf{Structural DoW Attack Demonstration.} We demonstrate structural DoW attacks on six MoE models under three LLM modes: \textbf{conversational, reasoning, and agentic tasks}. We further realize them with GBFA bit flips on four of these models. With only minimal parameter changes, our attacks consistently induce dramatic output inflation across all three modes, amplifying output token counts by up to 87$\times$ in the most severe cases.
\end{itemize}


\section{Background} \label{sec: background}

\subsection{Mixture-of-Experts}
MoE architectures scale language models efficiently by activating only a small subset of parameters per token, enabling trillion-parameter capacities at inference costs comparable to much smaller dense models. An MoE layer replaces the dense feed-forward block with a set of expert networks and a router that selects $k$ experts per token. Given a hidden state $\mathbf{h} \in \mathbb{R}^d$, the router computes
\begin{equation}
\mathbf{z} = \mathbf{h}^\top \mathbf{W}_g,\qquad
\mathbf{g} = \text{Softmax}(\text{TopK}(\mathbf{z}, k)),
\end{equation}
where $\mathbf{W}_g\!\in\!\mathbb{R}^{d\times N}$ is the router matrix. The layer output aggregates the selected experts:
\begin{equation}
\mathbf{y} = \sum_{i=1}^{N} g_i\, E_i(\mathbf{h}).
\end{equation}
Several recent MoE designs (e.g., DeepSeek-V3, Qwen3-MoE) further introduce \emph{shared experts} that are always activated for every token alongside the top-$k$ routed ones, capturing common knowledge while routed experts specialize.

MoE has become the dominant approach for scaling LLMs: Switch Transformer~\cite{fedus2022switchtransformersscalingtrillion} first pushed parameter counts to the trillion scale, Mixtral-8x7B~\cite{jiang2024mixtralexperts} reached GPT-3.5-level performance with far fewer active parameters, and recent systems such as DeepSeek-V2~\cite{deepseek_v2}, Qwen3~\cite{qwen3technicalreport}, and GPT-OSS-20B~\cite{openai2025gptoss120bgptoss20bmodel} continue to refine the architecture across reasoning, coding, and agentic workloads. Their growing adoption highlights the importance of understanding MoE-specific security vulnerabilities.

\subsection{Bit Flip Attacks}

Bit flip attacks (BFAs) exploit the fact that small, carefully chosen changes at the bit level of stored weights can cause disproportionately large shifts in model behavior while modifying only a tiny fraction of parameters~\cite{Rakin2019BFA, liu2017faultinjection, ghavami2022bdfablinddataadversarial, rakin2022tbfa}. This sensitivity stems from the underlying numerical representation. In IEEE 754 FP32, each value consists of 1 sign bit, 8 exponent bits, and 23 mantissa bits, and flipping bits in different fields produces qualitatively different perturbation scales: a sign-bit flip on $1.5$ yields $-1.5$, flipping the highest exponent bit turns $1.5$ into $\sim$$3.1\times10^{38}$, while flipping the lowest mantissa bit changes it by only $\sim$$10^{-7}$. Quantized INT8 weights show analogous behavior, where MSB flips cause large sign or magnitude jumps and LSB flips induce only marginal adjustments.

\textbf{Rowhammer} is the most widely studied software-based fault injection primitive for realizing such bit flips on commodity hardware~\cite{kim2014rowhammer}: by repeatedly accessing selected DRAM rows, an attacker induces flips in adjacent rows without physical access to the victim machine. End-to-end Rowhammer exploits have been demonstrated on both CPUs~\cite{kwong2020rambleed, seaborn2015exploiting, jattke2024zenhammer, frigo2020trrespass, jattke2022blacksmith, half-double, hassan2021utrr, HBM2_RH_Profile} and GPUs~\cite{Chris2025GPUHammer}, and have been used against neural networks for accuracy degradation~\cite{Rakin2019BFA, yao2020deephammer}, targeted misclassification~\cite{rakin2022tbfa, bai2023versatile, cai2021nmtstroke}, backdoor insertion~\cite{rakin2020tbt, zheng2023trojvit, cai2024deepvenom, chen2021proflip, DNN_Backdoor_dsn23, li2025oneflip}, and model extraction~\cite{rakin2022deepsteal}, establishing it as a mature and realistic threat vector. We analyze the router layers of various MoE-based LLMs and develop corresponding bit-flip strategies in Section~\ref{subsec:B-flip Attack on MoE Router}.

\section{Groundhog Bit-Flip Attack} \label{sec: Method}
\begin{figure*}[t]
  \centering
  \includegraphics[width=0.9\textwidth]{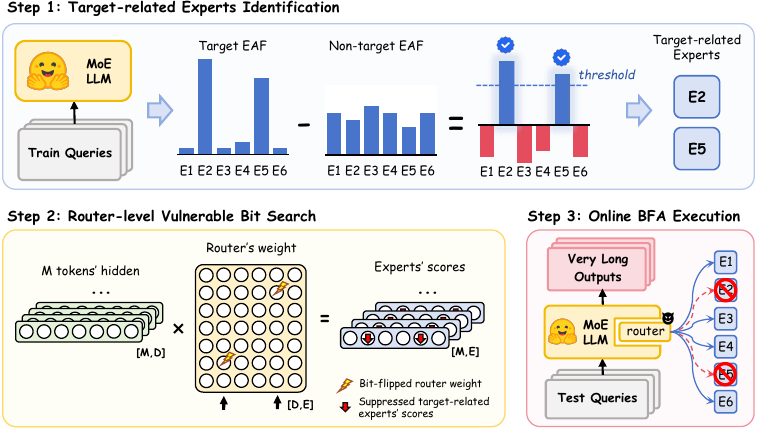}
  \caption{
Overview of our three-step GBFA pipeline.
\textbf{Step 1:} Identify target-related experts by comparing target-token EAF with non-target EAF and selecting experts above a threshold.
\textbf{Step 2:} Search for vulnerable router bits by evaluating whether flipping each candidate bit suppresses the scores of the selected target-related experts using cached hidden states.
\textbf{Step 3:} Execute the online bit-flip attack to suppress EOS generation and produce very long outputs.
}

  \label{fig:overview}
\end{figure*}

Figure~\ref{fig:overview} illustrates our attack pipeline, which addresses each research question in three steps. Full algorithm details are provided in Appendix~\ref{app:algorithm}.

To conduct GBFA against MoE LLMs, we raise three research questions:

\begin{itemize}[leftmargin=*]
    \item \textbf{RQ1:} Do MoE-based LLMs exhibit expert specialization for termination-related tokens, such as EOS and EOT?
    \item \textbf{RQ2:} Can we identify termination-related experts whose activation strongly influences response length?
    \item \textbf{RQ3:} Can targeted bit flip attacks on vulnerable experts effectively induce output inflation in MoE-based LLMs?
\end{itemize}

\subsection{Threat Model}

\noindent \textbf{Attack Scenarios.} We consider an on-cloud inference scenario where a service provider deploys an open-source MoE LLM for pay-per-token inference~\cite{deepseek_pricing}.
Users submit queries and are charged based on both input and output tokens. 
The adversary is hereby modeled as an unprivileged tenant co-located with the victim LLM on a shared physical server (e.g., in a public MLaaS environment), where workloads from multiple tenants share GPU resources in a time-sliced manner, consistent with prior work~\cite{Chris2025GPUHammer}. The adversary exploits hardware vulnerabilities (e.g., Rowhammer~\cite{kim2014rowhammer} or voltage glitching~\cite{6623558}) to inject bit flips into the victim's model parameters, requiring no elevated privileges or physical access.
The attacker aims to inflate output token consumption through strategic bit flips in the MoE router, causing a \textbf{DoW attack}, while maintaining semantic coherence to avoid detection.


\noindent \textbf{Attacker Capabilities.}
We assume white-box access to the victim MoE model, including the model hyper-parameters (e.g., number of experts) and parameters (e.g., router weights).
This is realistic because many MoE LLMs are released as public checkpoints with open weights, so the attacker can obtain the parameters directly.
We also assume the attacker is a co-located tenant that shares hardware with the victim (e.g., in multi-tenant MLaaS). The attacker can thus inject targeted bit flips into the victim LLM through fault injection techniques during model inference.

\noindent \textbf{Attack Objectives.}
The adversary aims to conduct a DoW attack (increased user's query cost to the LLM API) by controlling target token generation to: (1) suppress target token for long outputs, or (2) alter targeted token frequencies.

\noindent \textbf{Stealthiness Constraints.}
To ensure the stealthiness of GBFA, the attacker maintains task performance within acceptable thresholds, keeping original semantic coherence (e.g., perplexity, benchmark scores) largely unaffected.

\subsection{Target-related  Expert Activation Probability Modeling}\label{subsec:Target-related  Expert Activation Probability Modeling}

As shown in Step~1 in Figure~\ref{fig:overview}, we analyze the model's \textbf{Expert Activation Frequency (EAF)} at target versus non-target token positions. 

\noindent \textbf{Notation.} 
Consider an MoE model with $L$ MoE-FFN layers indexed by $l \in \{1, \ldots, L\}$, each containing $K$ experts. 
Let $\mathcal{S} = \{s_1, \ldots, s_N\}$ be a sample set in which the last token of every sequence is the target token , and let $|s|$ denote the length of sequence $s$. 
At token position $j$ of sequence $s$, let $\mathbb{1}_{l,i}^{s,j} \in \{0,1\}$ indicate whether expert $i$ in layer $l$ is activated. For models with shared experts, let $g_{l,i}^{s,j} \in \mathbb{R}_{\geq 0}$ denote the gate value of shared expert $i$ in layer $l$ at position $j$ of sequence $s$.

\noindent \textbf{Objective.} 
Our hypothesis is that \emph{if certain experts are responsible for sequence termination, then their activation patterns should differ sharply when generating the target token.} 
To quantify this effect, we aim to measure each expert's \emph{preference} for target token decoding over standard-token decoding.


\noindent \textbf{Estimator.} 
For expert $i$ in layer $l$, we define the \textbf{Target Activation Shift} $\tau_{l,i}$ as the difference between its EAF at target-token positions and its EAF at non-target positions:
\begin{equation}
\tau_{l,i}
=
\underbrace{\tfrac{1}{N}\!\sum_{s} \mathbb{1}_{l,i}^{s,|s|}}_{\text{target EAF}}
-
\underbrace{\tfrac{\sum_{s}\sum_{j<|s|} \mathbb{1}_{l,i}^{s,j}}{\sum_{s}(|s|-1)}}_{\text{non-target EAF}}.
\label{eq:routed experts}
\end{equation}
A high $\tau_{l,i}$ identifies an expert that rarely fires on standard tokens but frequently fires when emitting the target token. 
Since shared experts are always activated, the binary indicator $\mathbb{1}_{l,i}^{s,j}$ is uninformative for them. We instead track the shift in their gate magnitude. For shared expert $i$ in layer $l$, we define the \textbf{Target Gate Shift} $\Delta g_{l,i}$ analogously to $\tau_{l,i}$:
\begin{equation}
\Delta g_{l,i}
=
\underbrace{\tfrac{1}{N}\!\sum_{s} g_{l,i}^{s,|s|}}_{\text{target gate}}
-
\underbrace{\tfrac{\sum_{s}\sum_{j<|s|} g_{l,i}^{s,j}}{\sum_{s}(|s|-1)}}_{\text{non-target gate}}.
\label{eq:shared experts}
\end{equation}
A high $\Delta g_{l,i}$ identifies a shared expert whose contribution is sharply elevated when emitting the target token, marking it as a target-related shared expert.

We refer to such experts as \textbf{target-related experts} (e.g., EOS-related experts when the target is EOS), as their activation patterns directly encode termination behavior.

These findings answer \textbf{RQ1}: \textit{MoE LLMs exhibit strong expert specialization for target token generation, with a small subset of experts disproportionately responsible for sequence termination.}

\subsection{Target Expert Detection}

To identify such experts efficiently, we introduce two complementary detection methods—\textbf{Global Experts Detection} and \textbf{Local Experts Detection}—corresponding to Step~1 in Figure~\ref{fig:overview} and addressing \textbf{RQ2}.

\noindent\textbf{Global Experts Detection (GLOBAL).}
Global ranks all experts across all layers by  $\tau_{l,i}$ and $\Delta g_{l,i}$. Instead of selecting a fixed top-$k$, we label as target-related those experts whose scores exceed a threshold, ensuring the selected set corresponds to meaningful target token specialization. These experts are subsequently \textit{deactivated} to suppress target token routing and extend output length.

\noindent\textbf{Local Experts Detection (LOCAL).}
Local evaluates each layer independently to identify the layer that contributes most strongly to target-tokens control. 
For each layer $l$, we:
(1) select the top-$k$ experts within that layer according to $\tau_{l,i}$ and $\Delta g_{l,i}$, where $k$ equals the model’s native MoE routing top-$k$;
(2) deactivate only these experts in layer $l$;
(3) measure the resulting change in output length.
The layer producing the largest increase is denoted as $l^*$, indicating that it contains the most influential target-related experts. 
This procedure corresponds to the layer-level search illustrated in Figure~\ref{fig:overview}-Step~2.

\noindent\textbf{Comparing GLOBAL and LOCAL.}
GLOBAL incurs minimal computation but may modify more experts to achieve desired behavior manipulation. LOCAL is more precise, often leveraging fewer expert modifications and thus preserving good model utility, but requires $O(L)$ evaluations. Additionally, some MoE architectures (e.g., GPT-OSS~\cite{openai2025gptoss120bgptoss20bmodel}) distribute EOS behavior across multiple layers, making LOCAL less effective on those models.

Both GLOBAL and LOCAL remain stable even with very small sample sizes in the detection process (e.g., $N=10$), consistently identifying target-related experts. Together, they offer flexible trade-offs between stealthiness and computational cost.

\subsection{Bit-Flip Attack on MoE Router}\label{subsec:B-flip Attack on MoE Router}

Building on the identified target-related experts, we design a router-level bit-flip attack that directly alters their activation patterns (Step~3 in Figure~\ref{fig:overview}), addressing \textbf{RQ3}. A standard MoE router computes logits as
\begin{equation}
    s = h^\top W + b,
\end{equation}
where $h\!\in\!\mathbb{R}^D$, $W\!\in\!\mathbb{R}^{D\times E}$, and $b$ is optional. Among the evaluated models, Mixtral-8x7B, Phi-3.5-MoE, DeepSeek-V2-Lite, and Qwen3 use bias-free routers, so we flip bits in the expert columns $v_i$ of $W$; GPT-OSS-20B includes non-zero biases, so we flip bits in $b_i$ instead. Both strategies suppress target-related experts analogously to manual deactivation.

Full-model BFA~\cite{Rakin2019BFA} is infeasible at billion-parameter MoE scale, as every candidate bit requires a full forward pass. To avoid this cost, we propose an inference-free router-centric bit search. We run the model once on $N$ queries, cache the hidden states at target layer $\ell^*$ as an $[M,D]$ tensor, and reuse them for all bit evaluations. For each target-related expert $i$ with routing vector $v_i$, we flip every bit once, recompute routing logits from cached states, and measure the resulting expert activation frequency (EAF). We define the bit-flip effectiveness as
\begin{equation}
    \mathrm{BF\text{-}eff}(b)=
    \frac{\mathrm{EAF}_i^{\text{orig}}-\mathrm{EAF}_i^{(b)}}{\mathrm{EAF}_i^{\text{orig}}},
\end{equation}
and select the top-$B$ bits per expert. Flipping these bits produces a perturbed routing vector $\tilde{v}_i$ that reliably suppresses expert $i$ from the router's top-$k$, realizing \textbf{GBFA} without repeated inference.

\subsection{Online BFA Execution}

With the target bits identified, the attacker flips them in the deployed model's DRAM via Rowhammer~\cite{kim2014rowhammer, Chris2025GPUHammer} (Step~3 in Figure~\ref{fig:overview}). Since the perturbation modifies persistent router weights, a single successful attack suppresses target-related experts across all subsequent queries without further interaction.







\section{Experiment} \label{sec: experiment}


\subsection{Experiment Setup} \label{subsec: Experiment Setups}

\noindent \textbf{Models.} 
We evaluate six open-source MoE-based LLMs with diverse architectures (see Table~\ref{tab:moe_params} for specifications): Mixtral-8x7B-Instruct-v0.1~\cite{jiang2024mixtralexperts} (Mixtral), Phi-3.5-MoE-Instruct~\cite{phi3_techreport} (Phi-3.5), DeepSeek-V2-Lite-Chat~\cite{deepseek_v2} (DeepSeek), Qwen3-30B-A3B~\cite{qwen3technicalreport} (Qwen3), Qwen3-Coder-Next~\cite{qwen_qwen3_coder_next_tech_report} (Qwen3-Coder), and GPT-OSS-20B~\cite{openai2025gptoss120bgptoss20bmodel} (GPT-OSS). These models represent different design philosophies in MoE architectures, from efficient small-scale routing to fine-grained expert decomposition.

\begin{table}[h]
\centering
\scriptsize
\setlength{\tabcolsep}{5pt}
\caption{Overview of evaluated MoE architectures, showing total parameters, activated parameters per token, number of experts per layer, activated experts, and model depth.}
\label{tab:moe_params}
\resizebox{\linewidth}{!}{
\begin{tabular}{lccccc}
\toprule
\textbf{Model} & \textbf{\#Params (B)} & \textbf{Active (B)} & \textbf{\#Experts} & \textbf{\#Active} & \textbf{\#Layers} \\
\midrule
Mixtral      & 46.7 & 12.9 & 8   & 2  & 32 \\
Phi-3.5      & 41.9 &  6.6 & 16  & 2  & 32 \\
DeepSeek     & 15.7 &  2.4 & 64  & 6  & 27 \\
Qwen3        & 30.5 &  3.3 & 128 & 8  & 48 \\
Qwen3-Coder  & 80.0 &  3.0 & 512 & 10 & 48 \\
GPT-OSS      & 20.0 &  4.0 & 32  & 4  & 40 \\
\bottomrule
\end{tabular}
}
\vspace{-10pt} 
\end{table}

\noindent\textbf{Metrics.}
We use two metrics. For \textbf{attack effectiveness}, we compute the \textbf{Percentage of Token Generation Change (P)}:
\begin{equation}
\text{P}_i = \frac{T_{\text{deact}}^{(i)} - T_{\text{base}}^{(i)}}{T_{\text{base}}^{(i)}} \times 100\%.
\label{equ:P}
\end{equation}
$T_{\text{base}}^{(i)}$ and $T_{\text{deact}}^{(i)}$ denote baseline and attacked token counts, and the mean $\widehat{\text{P}}$ measures overall length increase. Figure~\ref{fig:ged_boundary_agnews_monitor} reports the underlying absolute token counts and their spread across blocked-expert counts. For \textbf{model utility}, we report Clean Accuracy (CA) on classification datasets, ROUGE-1 on Samsum, F1 on SQuAD, and \textbf{plan steps} on agent coding tasks.

\noindent \textbf{Datasets.}
We use the \textit{Stanford Alpaca} instruction-tuning dataset~\cite{alpaca} as the source of sample sequences for identifying target-related experts. To evaluate our attack, we use four representative datasets: \textit{AGNews}~\cite{NIPS2015_250cf8b5} and \textit{SST-2}~\cite{socher2013recursive} for classification, and \textit{Samsum}~\cite{gliwa2019samsum} and \textit{SQuAD 2.0}~\cite{rajpurkar-etal-2018-know} for generation. We follow prior work~\cite{wang2025badmoebackdooringmixtureofexpertsllms} to adopt system prompts for all downstream tasks. All models and datasets used in our experiments are publicly available. We use them for research purposes in accordance with their corresponding licenses and terms of use.

\noindent \textbf{Configuration.}
We target EOS. GLOBAL selects the top-$k_g$ EOS-related experts ranked by frequency difference, where $k_g$ varies per model to balance attack effectiveness and efficiency: 37 for Mixtral, 16 for Phi-3.5, 40 for DeepSeek, and 25 for GPT-OSS. LOCAL follows each model's default router top-$k$ (Table~\ref{tab:moe_params}). We use greedy decoding and set max tokens to 1024. The attack does not depend on the decoding strategy: Appendix~\ref{app:decoding} shows that the length inflation persists under temperature and nucleus sampling and under a repetition penalty, and Appendix~\ref{app:budget} reports the effect of the token budget and motivates our moderate choice of 1024.

\noindent\textbf{Platform.} Experiments were conducted on three NVIDIA A100 80GB GPUs. For Section \ref{subsec:Target-related  Expert Activation Probability Modeling}, we computed the $\tau_{l,i}$ or $\Delta g_{l,i}$ over 1,000 samples.  For identifying  attacked bits in the bit-flip attack pipeline, the average cost was \textbf{16.4 s per expert}.

\subsection{Manual EOS-related Experts Deactivation} \label{sub:Manual experts deactivation}

To validate that \textbf{EOS-related experts} control output length, we manually deactivate the identified experts by setting their logits to \textbf{-inf}, establishing upper bounds for bit-flip attack effectiveness. Table~\ref{tab:Comparison_manual} reports $P$ across 100 samples per dataset under LOCAL, GLOBAL, and a Random-deactivation baseline (50 random experts).

Both LOCAL and GLOBAL substantially amplify output, consistently outperforming Random by orders of magnitude, validating our expert-detection procedure. The relative strength of LOCAL vs.\ GLOBAL reveals how termination is implemented in each model. When EOS is controlled by a few concentrated experts, LOCAL suffices and can even exceed GLOBAL (e.g., DeepSeek AGNews: $3.81\times10^{4}\%$ vs.\ $226\%$). When termination relies on an inter-layer path, GLOBAL becomes necessary (e.g., GPT-OSS AGNews: $556\%$ vs.\ $15.9\%$). Classification tasks yield larger $P$ than generation tasks, as short baseline outputs amplify into thousand-fold increases.

\begin{figure*}[ht]
  \centering
  \includegraphics[width=\textwidth]{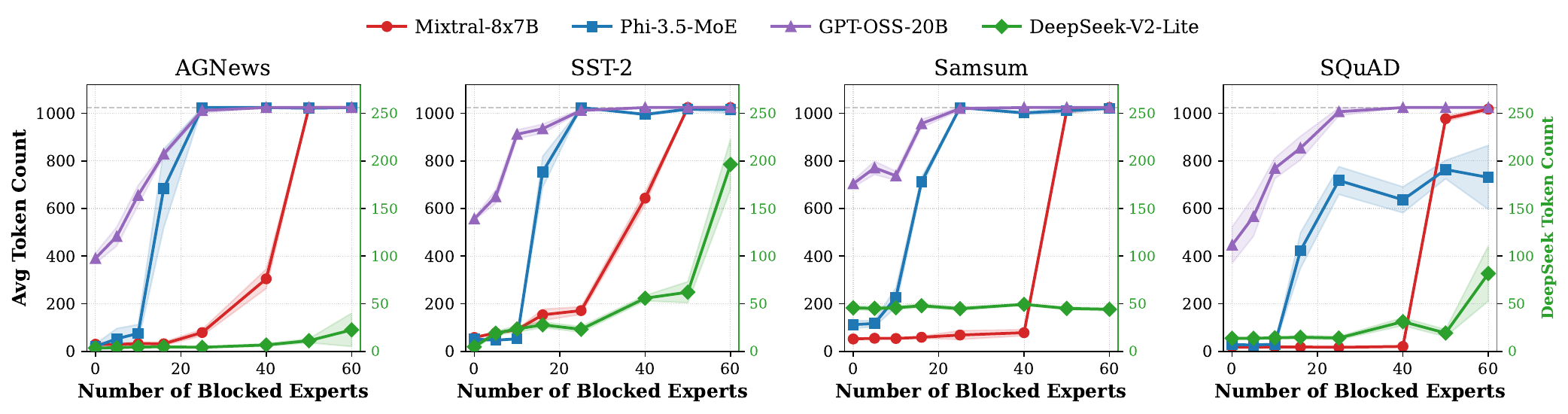}
  \caption{Average output token count under GLOBAL manual deactivation as a function of blocked experts. The dashed line marks \texttt{max\_new\_tokens}; DeepSeek-V2-Lite uses a separate right axis.}
  \label{fig:ged_boundary_agnews_monitor}
\end{figure*}

Figure~\ref{fig:ged_boundary_agnews_monitor} shows that output token count generally increases with more blocked experts. GPT-OSS and Phi-3.5 quickly saturate \texttt{max\_new\_tokens} and plateau. DeepSeek shows irregular behavior on Samsum, likely due to the model's strong prior on short responses for this dataset.

Without EOS-related experts, models generate repetitive but coherent text loops that preserve linguistic quality while losing termination control (qualitative examples on Mixtral in Appendix~\ref{app:attacked_output}). This output amplification potential motivates our bit-flip targets.

\begin{table}[h]
\centering
\scriptsize
\setlength{\tabcolsep}{4pt}
\renewcommand{\arraystretch}{0.95}
\caption{$P$ across all samples for manual expert deactivation (LOCAL vs.\ GLOBAL), compared against a random-deactivation baseline.}
\label{tab:Comparison_manual}
\resizebox{\linewidth}{!}{
\begin{tabular}{l|l|c|c|c|c}
\toprule
Model & Method (\#exp) & AGNews (\%) & SST-2 (\%) & Samsum (\%) & SQuAD (\%) \\
\midrule
\multirow{3}{*}{Mixtral}
& Random \hfill(50)  & 685                         & 188                         & 149                         & -9.28 \\
& LOCAL  \hfill(2)   & $2.18\times10^{3}$          & $2.23\times10^{3}$          & $2.19\times10^{3}$          & $3.64\times10^{3}$ \\
& GLOBAL \hfill(37)  & $\mathbf{2.74\times10^{3}}$ & $\mathbf{3.41\times10^{3}}$ & $\mathbf{389}$              & $\mathbf{52.9}$ \\
\midrule
\multirow{3}{*}{Phi-3.5}
& Random \hfill(50)  & 215                         & 1.14                        & 30.1                        & 155 \\
& LOCAL  \hfill(2)   & $2.82\times10^{4}$          & $1.83\times10^{3}$          & 20.5                        & 217 \\
& GLOBAL \hfill(16)  & $\mathbf{2.99\times10^{4}}$ & $\mathbf{1.74\times10^{3}}$ & $\mathbf{1.27\times10^{3}}$ & $\mathbf{1.89\times10^{3}}$ \\
\midrule
\multirow{3}{*}{DeepSeek}
& Random \hfill(50)  & 67.9                        & 125                         & 5.97                        & 18.8 \\
& LOCAL  \hfill(6)   & $\mathbf{3.81\times10^{4}}$ & $5.82\times10^{3}$          & $\mathbf{78.6}$             & $\mathbf{1.00\times10^{4}}$ \\
& GLOBAL \hfill(40)  & 226 & $\mathbf{2.81\times10^{3}}$ & 14                         & 140 \\
\midrule
\multirow{3}{*}{GPT-OSS}
& Random \hfill(50)  & -3.43                       & 10.3                        & 11.6                        & 2.29 \\
& LOCAL  \hfill(4)   & 15.9                        & 17.3                        & 36.7                        & 9.30 \\
& GLOBAL \hfill(25)  & $\mathbf{556}$              & $\mathbf{468}$              & $\mathbf{177}$              & $\mathbf{445}$ \\
\bottomrule
\end{tabular}
}
\end{table}

\subsection{Groundhog Bit-Flip Attack}\label{subsec: Bit-flip Attack}

We implement GBFA targeting router weights to manipulate EOS-related expert activation. For each \textbf{EOS-related expert}, we identified three critical router weights using the method described in Section \ref{subsec:B-flip Attack on MoE Router}, with each weight requiring only a single bit, resulting in \textbf{three total bit modifications} per expert. For models with router bias parameters (e.g., GPT-OSS), we prioritized attacking these biases as they directly offset routing logits, providing more efficient manipulation. Table \ref{tab:Comparison_bflip} presents the P described in Equation \ref{equ:P} under bit-flip attacks using LOCAL and GLOBAL selection strategies.

The results show that \textbf{GBFA} achieves substantial output manipulation comparable to manual deactivation. Since attacking weight matrices involves complex inter-weight dependencies, where multiple weights collectively determine routing logits across varying input distributions, bit-flips produce approximate rather than exact expert suppression. This approximation can occasionally yield stronger deactivation effects than manual blocking, as the perturbations propagate through weight interactions in non-trivial ways. Overall, the results confirm that GBFA is a viable and effective attack: despite operating at the bit level with minimal parameter changes, it reliably induces significant output inflation across models and datasets.

Notably, GPT-OSS demonstrates near-manual performance, achieving comparable results to manual deactivation. This exceptional effectiveness stems from directly attacking bias parameters, which provide a linear offset to all routing decisions without inter-parameter interference. These findings validate that GBFA can successfully manipulate output length through EOS-expert targeting, with bias parameters offering optimal attack surfaces when available.






\begin{table}[h]
\centering
\scriptsize
\setlength{\tabcolsep}{4pt}
\renewcommand{\arraystretch}{0.9}
\caption{P across All Samples for Bit-flip attack (LOCAL vs GLOBAL).}
\label{tab:Comparison_bflip}
\resizebox{\linewidth}{!}{
\begin{tabular}{l|l|c|c|c|c}
\toprule
Model & Method (\#exp) & AGNews (\%) & SST-2 (\%) & Samsum (\%) & SQuAD (\%) \\
\midrule

\multirow{2}{*}{Mixtral}
  & LOCAL \hfill(2)  & 202        & 110        & 1.06       & 39.3 \\
  & GLOBAL \hfill(37)& \num{2.51e3} & \num{4.99e3} & \num{2.92e3} & 48.1 \\
  \midrule

  \multirow{2}{*}{Phi-3.5}
  & LOCAL \hfill(2)  & 526        & 42.1       & 148        & 45.3 \\
  & GLOBAL \hfill(16)& 131        & 20.8       & 222        & 167  \\

\midrule

\multirow{2}{*}{DeepSeek}
& LOCAL \hfill(6) & \num{3.84e4} & \num{8.73e4} & \num{1.61e3} & \num{4.05e2} \\
& GLOBAL \hfill(40) & \num{4.64e4} & \num{4.44e4} & \num{3.04e3} & \num{1.49e4} \\
\midrule

\multirow{2}{*}{GPT-OSS}
& LOCAL \hfill(4) & 15.9 & 17.3 & 36.7 & 9.30 \\
& GLOBAL \hfill(25) & 556       & 468            & 177     & 445 \\
\bottomrule
\end{tabular}
}
\end{table}

\subsection{Performance Impact of GBFA on MoE Models}\label{subsec: Performance comparison}

We evaluated model performance under \textbf{GBFA} across four MoE architectures using 1,000 randomly selected samples per dataset. Table \ref{tab:performance_comparation} presents the results comparing baseline, manual deactivation, and bit flip conditions.

Performance degradation correlates with attack effectiveness (measured by Equation \ref{equ:P}). For Mixtral, manual deactivation causes substantial drops in Clean Accuracy (CA) decreases from 0.838 baseline to 0.602 (LOCAL) and 0.459 (GLOBAL) on AGNews. In contrast, bit-flip attacks preserve performance better, with CA remaining at 0.840 (LOCAL) and 0.800 (GLOBAL). Interestingly, some configurations show performance improvements: Phi-3.5's ROUGE-1 on Samsum increases from 0.362 to 0.347-0.382 under LOCAL attacks. This aligns with recent findings~\cite{fayyaz2025steeringmoellmsexpert} that selective expert manipulation can enhance task-specific performance through improved alignment.

Perplexity (PPL) measurements confirm that most attacks maintain linguistic coherence. PPL values remain below 10 for most configurations, indicating meaningful text generation rather than gibberish. The notable exception is DeepSeek under GLOBAL bit-flip attacks, where PPL explodes to 91,839 (SST-2) and 5,288,426 (Samsum), signaling complete language generation breakdown. We attribute this to the flipped bits also perturbing router weights on which DeepSeek relies for coherent generation: corruption typically emerges within the first few generated tokens and then propagates, since early tokens condition all subsequent ones. Incorporating the perplexity of the first few tokens as an additional criterion in the bit search could mitigate this effect, which we leave for future work.

Overall, \textbf{GBFA} on MoE routers achieves effective output manipulation while maintaining relative stealth by preserving model quality metrics even as they dramatically extend output length.

\begin{table}[h]
\centering
\scriptsize
\setlength{\tabcolsep}{4pt}
\renewcommand{\arraystretch}{1.25}
\caption{Performance of different models on four datasets under manual and bit-flip expert manipulation.}
\label{tab:performance_comparation}
\resizebox{\linewidth}{!}{%
\begin{tabular}{l l cc cc cc cc}
\toprule
\textbf{Model} & \textbf{Method} 
& \multicolumn{2}{c}{\textbf{AGNews}} 
& \multicolumn{2}{c}{\textbf{SST-2}} 
& \multicolumn{2}{c}{\textbf{Samsum}} 
& \multicolumn{2}{c}{\textbf{SQuAD}} \\
\cmidrule(lr){3-4} \cmidrule(lr){5-6} \cmidrule(lr){7-8} \cmidrule(lr){9-10}
& & CA & PPL & CA & PPL & R-1 & PPL & F1 & PPL \\
\midrule
\multirow{5}{*}{DeepSeek}
& Baseline        & 0.810 & 1.35    & 0.919  & 1.69    & 0.364  & 1.38    & 0.362  & 1.22 \\
& LOCAL (manual)  & 0.970 & 1.17    & 0.793  & 2.33    & 0.270  & 1.49    & 0.070  & 1.40 \\
& GLOBAL (manual) & 0.395 & 1.82    & 0.836  & 1.73    & 0.205  & 1.60    & 0.159  & 1.82 \\
& LOCAL (GBFA)  & 0.380 & 1.82    & 0.020  & 1.81    & 0.117  & 1.36    & 0.021  & 1.21 \\
& GLOBAL (GBFA) & 0.160 & 4.4e3   & 0.000  & 9.2e4   & 0.025  & 5.3e6   & 0.081  & 5.1e6 \\
\midrule
\multirow{5}{*}{Mixtral}
& Baseline        & 0.838 & 1.40    & 0.809  & 1.47    & 0.362  & 1.39    & 0.253  & 1.13 \\
& LOCAL (manual)  & 0.602 & 2.37    & 0.698  & 3.73    & 0.171  & 1.27    & 0.071  & 1.51 \\
& GLOBAL (manual) & 0.459 & 2.52    & 0.508  & 3.91    & 0.086  & 1.18    & 0.058  & 1.43 \\
& LOCAL (GBFA)  & 0.840 & 1.33    & 0.890  & 1.47    & 0.102  & 1.91    & 0.237  & 1.09 \\
& GLOBAL (GBFA) & 0.800 & 1.89    & 0.920  & 1.68    & 0.252  & 1.66    & 0.242  & 1.23 \\
\midrule
\multirow{5}{*}{Phi-3.5}
& Baseline        & 0.856 & 1.13    & 0.862  & 1.25    & 0.375  & 1.21    & 0.235  & 1.10 \\
& LOCAL (manual)  & 0.569 & 2.26    & 0.681  & 1.63    & 0.347  & 1.29    & 0.096  & 1.36 \\
& GLOBAL (manual) & 0.076 & 6.12    & 0.011  & 8.95    & 0.025  & 1.46    & 0.052  & 2.20 \\
& LOCAL (GBFA)  & 0.800 & 1.18    & 0.830  & 1.33    & 0.382  & 1.23    & 0.263  & 1.07 \\
& GLOBAL (GBFA) & 0.750 & 1.98    & 0.840  & 1.33    & 0.339  & 1.46    & 0.135  & 1.26 \\
\midrule
\multirow{5}{*}{GPT-OSS}
& Baseline        & 0.800 & 1.36    & 0.940  & 1.37    & 0.199  & 1.36    & 0.049  & 1.12 \\
& LOCAL (manual)  & 0.800 & 1.37    & 0.940  & 1.41    & 0.177  & 1.37    & 0.044  & 1.12 \\
& GLOBAL (manual) & 0.790 & 1.56    & 0.880  & 1.61    & 0.065  & 1.33    & 0.045  & 1.17 \\
& LOCAL (GBFA)  & 0.800 & 1.37    & 0.940  & 1.41    & 0.177  & 1.37    & 0.044  & 1.12 \\
& GLOBAL (GBFA) & 0.790 & 1.56    & 0.880  & 1.61    & 0.065  & 1.33    & 0.045  & 1.17 \\
\bottomrule
\end{tabular}%
}
\end{table}

\section{Broader Attack Scenarios} \label{sec: Broader Attack Scenarios}

\subsection{Attack the Plan Mode in Agent Coding}\label{sec:agentic_coding}
We evaluate whether GBFA can prolong an LLM agent's planning behavior in long-horizon coding tasks. We use \textbf{Qwen3-Coder}, a model designed specifically for coding agents and local development, deployed under the \textbf{smolagents} \texttt{CodeAgent} framework, a Hugging Face library for easily running agents on Hugging Face models. The framework alternates between planning and action steps, and we target the chat-template end-of-turn token \texttt{<|im\_end|>}.

\textbf{Setup.} We design 10 self-contained Python repository sandboxes spanning multiple task categories (full descriptions in Appendix~\ref{app:sandbox_tasks}). Each sandbox is reset to a clean snapshot before every run, with completion verified by an associated unit-test suite. We compare a clean baseline against two manual deactivation configurations, \textbf{LOCAL} and \textbf{GLOBAL}, targeting the top-20 routed experts ranked by $\tau_{l,i}$ and the top-4 shared experts ranked by $\Delta g_{l,i}$. Each agent runs with a 25-step limit, planning every 3 steps, and 2048 max new tokens per generation.

\noindent\textbf{Plan-mode results.} Table~\ref{tab:plan_mode_block} shows that GLOBAL deactivation reliably amplifies plan-token consumption: all 10 sandboxes hit the maximum step budget, with mean $\widehat{P} = +501.5\%$. LOCAL remains inconsistent and statistically negligible. The amplification manifests as plan-output degeneration into syntactically valid character loops, illustrated on five sandboxes in Appendix~\ref{app:Attacked Output of Plan Mode}.

\begin{table}[t]
\centering
\footnotesize
\setlength{\tabcolsep}{3pt}
\renewcommand{\arraystretch}{1.2}
\caption{Plan-mode manual expert deactivation on Qwen3-Coder-Next across 10 coding sandboxes. We report plan steps (\#), average plan tokens, and the relative change $P$ in plan-token totals against the baseline.}
\label{tab:plan_mode_block}
\resizebox{\linewidth}{!}{%
\begin{tabular}{l cc cc r cc r}
\toprule
& \multicolumn{2}{c}{\textbf{Baseline}} & \multicolumn{3}{c}{\textbf{LOCAL (manual)}} & \multicolumn{3}{c}{\textbf{GLOBAL (manual)}} \\
\cmidrule(lr){2-3} \cmidrule(lr){4-6} \cmidrule(lr){7-9}
\textbf{Sandbox} & \# & tokens & \# & tokens & $P$ & \# & tokens & $P$ \\
\midrule
calculator\_refactor   & 4 & 540.2 & 3 & 888.0  & $+23.3\%$  & 9 & 1869.6 & $+678.6\%$ \\
bug\_fix               & 3 & 620.0 & 3 & 611.3  & $-1.4\%$   & 9 & 2048.0 & $+891.0\%$ \\
feature\_add           & 3 & 507.7 & 3 & 565.7  & $+11.4\%$  & 9 & 1231.7 & $+627.8\%$ \\
rename\_refactor       & 4 & 461.8 & 4 & 620.0  & $+34.3\%$  & 9 & 1095.1 & $+433.6\%$ \\
test\_coverage         & 7 & 509.1 & 4 & 607.8  & $-31.8\%$  & 9 & 1061.6 & $+168.1\%$ \\
multi\_file\_refactor  & 6 & 585.0 & 5 & 891.0  & $+26.9\%$  & 9 & 1622.1 & $+315.9\%$ \\
bug\_hunt              & 3 & 405.3 & 3 & 568.0  & $+40.1\%$  & 9 & 1528.9 & $+1031.6\%$ \\
performance\_optimize  & 5 & 508.8 & 3 & 690.0  & $-18.6\%$  & 9 & 1005.8 & $+255.8\%$ \\
error\_handling        & 4 & 451.5 & 3 & 555.7  & $-7.7\%$   & 9 & 1464.8 & $+630.0\%$ \\
dependency\_swap       & 3 & 602.3 & 3 & 658.7  & $+9.4\%$   & 9 & 1666.6 & $+730.0\%$ \\
\midrule
\textbf{Mean}          & 4.2 & 519.2 & 3.4 & 665.6 & $+8.6\%$ & 9.0 & 1459.4 & $+576.2\%$ \\
\bottomrule
\end{tabular}%
}
\end{table}

\subsection{Attack the   End-of-Thinking Token}
To evaluate whether our attack transfers to other types of tokens, we apply it to the thinking process of GPT-OSS and Qwen3. Following the same procedure as the Groundhog Bit-Flip Attack, we use LOCAL and GLOBAL to identify experts that are closely associated with the \textbf{end of thinking (EOT)} token and we manually deactivate these experts. The results are shown in Table \ref{tab:gptoss_led_ged}. We observe a clear increase in the number of tokens generated during the thinking phase. In the most extreme case, the model continues thinking until the token budget is exhausted without producing the expected answer. This experiment demonstrates that our attack also applies to other specific tokens.

\begin{table}[h]
    \centering
    \small
    \setlength{\tabcolsep}{5pt}
    \renewcommand{\arraystretch}{0.9}
    \caption{Results of GPT-OSS and Qwen3 under LOCAL and GLOBAL manual deactivation of EOT-related experts (percentage increase in thinking length).}
    \label{tab:gptoss_led_ged}
    \resizebox{\linewidth}{!}{
    \begin{tabular}{l|l|c|c|c|c}
    \toprule
    \textbf{Model} & \textbf{Method (manual)} & \textbf{AGNews (\%)} & \textbf{SST-2 (\%)} & \textbf{Samsum (\%)} & \textbf{SQuAD (\%)} \\
    \midrule
    \multirow{2}{*}{GPT-OSS} & LOCAL  & 19.7 & 15.2 & 58.4 & 25.9 \\
                            & GLOBAL & \num{7.15e2} & \num{1.27e3} & 365 & \num{1.33e3} \\
    \midrule
    \multirow{2}{*}{Qwen3}   & LOCAL  & 6.28 & 1.61 & 4.18 & -1.37 \\
                            & GLOBAL & 204 & 224 & 230 & 271 \\
    \bottomrule
    \end{tabular}
    }
    \end{table}

\section{Discussion} \label{sec: discussion}
\subsection{Feasibility under Distributed Inference}
Large MoE models are often served with tensor parallelism (TP) and pipeline parallelism (PP). We discuss whether GBFA still works when router parameters are split or replicated across devices.
Under PP, each layer and its router stay on one device. The router bits we target are not split, so the attack needs the same number of flips as in the single-device setting. PP thus does not increase the difficulty.
Under TP, the router weights may be replicated on several devices, which would require flipping the same bit on each replica. The attacker can avoid this by corrupting the shared page cache that the replicas load from, so that one poisoned copy reaches all devices. Prior fault-injection work uses the same idea~\cite{DNN_Backdoor_dsn23}.
Neither scheme therefore breaks the feasibility of GBFA.

\subsection{Potential Defenses}
Potential defenses against GBFA fall into two categories.

\noindent \textbf{Software-level mitigations} such as API-level output length caps~\cite{openai_api, anthropic_api} reduce per-request damage but cannot prevent the underlying attack. Output-pattern detectors such as Breaking the Loop~\cite{yu2025breakingloop} can interrupt loop behavior in real time, but legitimate repetitive outputs trigger the same signal, limiting practical reliability. Neither class of defense remediates the root cause.



\noindent\textbf{System-level protections} such as ECC~\cite{sridharan2015memory} or TEE-based isolation~\cite{sun2023shadownet, gnnvault} can detect accidental faults but provide limited security against adversarial bit-flips and incur nontrivial overhead.

These gaps highlight the need for \textit{MoE-specific defenses} that constrain routing sensitivity and verify parameter integrity.

\section{Conclusion} \label{sec: Conclusion}
In this paper, we proposed \textbf{Groundhog Bit-Flip Attack (GBFA)}, a structural Denial-of-Wallet attack that exploits MoE routing through targeted router-level bit-flips. Our analysis revealed strong target-related expert specialization, enabling attackers to suppress termination-critical experts by flipping only a few bits. Using GBFA, we efficiently identified vulnerable experts and showed that these small perturbations can induce massive output-length inflation while largely preserving model utility. Experiments across six MoE LLMs and three LLM modes  confirm that routing-layer corruption constitutes a previously overlooked vulnerability. These findings underscore the need for MoE-specific defenses to safeguard router integrity and prevent targeted manipulation of expert activation.

\section{Limitations} \label{sec: Limitations}
\begin{itemize}[leftmargin=*]
    \item \textbf{Strong threat model.} GBFA assumes white-box access to the victim MoE model and the ability to inject targeted bit flips into selected router parameters. Although this follows prior bit-flip attack settings, it may not hold for all practical MLaaS with strong isolation or integrity checks.
    \item \textbf{System-level cost analysis.} We quantify vulnerable bit search cost and model-level attack effectiveness, but the full end-to-end attack cost also depends on hardware-specific factors such as memory placement and BFA reliability.
    \item \textbf{No end-to-end hardware demonstration.} We identify vulnerable bits and measure the output inflation once they are flipped, but we do not perform an end-to-end hardware exploit that physically flips them on a live system. System-level defenses such as ECC memory or TEE-based isolation raise the difficulty of such flips. However, they do not fully prevent adversarial bit flips and add nontrivial overhead, so they mitigate rather than eliminate the threat.
    \item \textbf{Theoretical analysis.} Our study is mainly empirical. We show that EOS and EOT behavior can concentrate in a small set of MoE experts, but we do not provide a formal theoretical explanation of how routing logits, expert specialization, and output-length inflation are related.
\end{itemize}

\section*{Acknowledgments}
Portions of this research were conducted with high performance computational resources provided by Louisiana State University (\url{http://www.hpc.lsu.edu}) and the Louisiana Optical Network Infrastructure (\url{http://www.loni.org}).

\bibliography{ref}

@inproceedings {Chris2025GPUHammer,
author = {Chris S. Lin and Joyce Qu and Gururaj Saileshwar},
title = {{GPUHammer}: Rowhammer Attacks on {GPU} Memories are Practical},
booktitle = {34th USENIX Security Symposium (USENIX Security 25)},
year = {2025},
isbn = {978-1-939133-52-6},
address = {Seattle, WA},
pages = {5719--5738},
url = {https://www.usenix.org/conference/usenixsecurity25/presentation/lin-shaopeng},
publisher = {USENIX Association},
month = aug
}

@inproceedings{Rakin2019BFA,
  author    = {Adnan Siraj Rakin and Zhezhi He and Deliang Fan},
  title     = {Bit-Flip Attack: Crushing Neural Network with Progressive Bit Search},
  booktitle = {Proceedings of the IEEE International Conference on Computer Vision (ICCV)},
  pages     = {1211--1220},
  year      = {2019}
}

@misc{alpaca,
  author = {Rohan Taori and Ishaan Gulrajani and Tianyi Zhang and Yann Dubois and Xuechen Li and Carlos Guestrin and Percy Liang and Tatsunori B. Hashimoto },
  title = {Stanford Alpaca: An Instruction-following LLaMA model},
  year = {2023},
  publisher = {GitHub},
  journal = {GitHub repository},
}

@article{ding2025moecho,
  title={MoEcho: Exploiting Side-Channel Attacks to Compromise User Privacy in Mixture-of-Experts LLMs},
  author={Ding, Ruyi and Xu, Tianhong and Shen, Xinyi and Ding, Aidong Adam and Fei, Yunsi},
  journal={arXiv preprint arXiv:2508.15036},
  year={2025}
}

@misc{deepseek_pricing,
  author       = {DeepSeek},
  title        = {DeepSeek API Documentation: Models and Pricing},
  howpublished = {\url{https://api-docs.deepseek.com/quick_start/pricing}},
  year         = {2025}
}

@article{jiang2024mixtralexperts,
  title={Mixtral of Experts},
  author={Jiang, Albert Q. and Sablayrolles, Alexandre and Roux, Antoine and Mensch, Arthur and Savary, Blanche and Bamford, Chris and Chaplot, Devendra Singh and Casas, Diego de las and Hanna, Emma Bou and Bressand, Florian and others},
  journal={arXiv preprint arXiv:2401.04088},
  year={2024}
}

@misc{phi3_techreport,
  author       = {Abdin, Marah et al.},
  title        = {Phi-3 Technical Report: A Highly Capable Language Model Locally on Your Phone},
  year         = {2024},
  eprint       = {2404.14219},
  archivePrefix= {arXiv},
  primaryClass = {cs.CL},
  url          = {https://arxiv.org/abs/2404.14219}
}

@misc{deepseek_v2,
      title={DeepSeek-V2: A Strong, Economical, and Efficient Mixture-of-Experts Language Model}, 
      author={DeepSeek-AI},
      year={2024},
      eprint={2405.04434},
      archivePrefix={arXiv},
      primaryClass={cs.CL}
}

@misc{openai2025gptoss120bgptoss20bmodel,
      title={gpt-oss-120b \& gpt-oss-20b Model Card}, 
      author={OpenAI},
      year={2025},
      eprint={2508.10925},
      archivePrefix={arXiv},
      primaryClass={cs.CL},
      url={https://arxiv.org/abs/2508.10925}, 
}

@misc{fayyaz2025steeringmoellmsexpert,
      title={Steering MoE LLMs via Expert (De)Activation}, 
      author={Mohsen Fayyaz and Ali Modarressi and Hanieh Deilamsalehy and Franck Dernoncourt and Ryan Rossi and Trung Bui and Hinrich Schütze and Nanyun Peng},
      year={2025},
      eprint={2509.09660},
      archivePrefix={arXiv},
      primaryClass={cs.CL},
      url={https://arxiv.org/abs/2509.09660}, 
}

@misc{fedus2022switchtransformersscalingtrillion,
      title={Switch Transformers: Scaling to Trillion Parameter Models with Simple and Efficient Sparsity}, 
      author={William Fedus and Barret Zoph and Noam Shazeer},
      year={2022},
      eprint={2101.03961},
      archivePrefix={arXiv},
      primaryClass={cs.LG},
      url={https://arxiv.org/abs/2101.03961}, 
}

@inproceedings{NIPS2015_250cf8b5,
 author = {Zhang, Xiang and Zhao, Junbo and LeCun, Yann},
 booktitle = {Advances in Neural Information Processing Systems},
 editor = {C. Cortes and N. Lawrence and D. Lee and M. Sugiyama and R. Garnett},
 pages = {},
 publisher = {Curran Associates, Inc.},
 title = {Character-level Convolutional Networks for Text Classification},
 url = {https://proceedings.neurips.cc/paper_files/paper/2015/file/250cf8b51c773f3f8dc8b4be867a9a02-Paper.pdf},
 volume = {28},
 year = {2015}
}

@inproceedings{socher2013recursive,
  title={Recursive deep models for semantic compositionality over a sentiment treebank},
  author={Socher, Richard and Perelygin, Alex and Wu, Jean and Chuang, Jason and Manning, Christopher D and Ng, Andrew Y and Potts, Christopher},
  booktitle={Proceedings of the 2013 conference on empirical methods in natural language processing},
  pages={1631--1642},
  year={2013}
}

@article{gliwa2019samsum,
  title={SAMSum corpus: A human-annotated dialogue dataset for abstractive summarization},
  author={Gliwa, Bogdan and Mochol, Iwona and Biesek, Maciej and Wawer, Aleksander},
  journal={arXiv preprint arXiv:1911.12237},
  year={2019}
}

@inproceedings{rajpurkar-etal-2018-know,
    title = "Know What You Don{'}t Know: Unanswerable Questions for {SQ}u{AD}",
    author = "Rajpurkar, Pranav  and
      Jia, Robin  and
      Liang, Percy",
    editor = "Gurevych, Iryna  and
      Miyao, Yusuke",
    booktitle = "Proceedings of the 56th Annual Meeting of the Association for Computational Linguistics (Volume 2: Short Papers)",
    month = jul,
    year = "2018",
    address = "Melbourne, Australia",
    publisher = "Association for Computational Linguistics",
    url = "https://aclanthology.org/P18-2124",
    doi = "10.18653/v1/P18-2124",
    pages = "784--789",
    eprint={1806.03822},
    archivePrefix={arXiv},
    primaryClass={cs.CL}
}

@INPROCEEDINGS{6623558,
  author={Moro, Nicolas and Dehbaoui, Amine and Heydemann, Karine and Robisson, Bruno and Encrenaz, Emmanuelle},
  booktitle={2013 Workshop on Fault Diagnosis and Tolerance in Cryptography}, 
  title={Electromagnetic Fault Injection: Towards a Fault Model on a 32-bit Microcontroller}, 
  year={2013},
  volume={},
  number={},
  pages={77-88},
  doi={10.1109/FDTC.2013.9}}

@misc{wang2025badmoebackdooringmixtureofexpertsllms,
      title={BadMoE: Backdooring Mixture-of-Experts LLMs via Optimizing Routing Triggers and Infecting Dormant Experts}, 
      author={Qingyue Wang and Qi Pang and Xixun Lin and Shuai Wang and Daoyuan Wu},
      year={2025},
      eprint={2504.18598},
      archivePrefix={arXiv},
      primaryClass={cs.CR},
      url={https://arxiv.org/abs/2504.18598}, 
}

@INPROCEEDINGS{liu2017faultinjection,
  author={Liu, Yannan and Wei, Lingxiao and Luo, Bo and Xu, Qiang},
  booktitle={2017 IEEE/ACM International Conference on Computer-Aided Design (ICCAD)}, 
  title={Fault injection attack on deep neural network}, 
  year={2017},
  volume={},
  number={},
  pages={131-138},
  doi={10.1109/ICCAD.2017.8203770}}

@misc{ghavami2022bdfablinddataadversarial,
      title={BDFA: A Blind Data Adversarial Bit-flip Attack on Deep Neural Networks}, 
      author={Behnam Ghavami and Mani Sadati and Mohammad Shahidzadeh and Zhenman Fang and Lesley Shannon},
      year={2022},
      eprint={2112.03477},
      archivePrefix={arXiv},
      primaryClass={cs.CR},
      url={https://arxiv.org/abs/2112.03477}, 
}

@INPROCEEDINGS{gnnvault,
  author={Ding, Ruyi and Xu, Tianhong and Ding, Aidong Adam and Fei, Yunsi},
  booktitle={2025 62nd ACM/IEEE Design Automation Conference (DAC)}, 
  title={Graph in the Vault: Protecting Edge GNN Inference with Trusted Execution Environment}, 
  year={2025},
  volume={},
  number={},
  pages={1-7},
  doi={10.1109/DAC63849.2025.11132467}}

@inproceedings{sun2023shadownet,
  title={Shadownet: A secure and efficient on-device model inference system for convolutional neural networks},
  author={Sun, Zhichuang and Sun, Ruimin and Liu, Changming and Chowdhury, Amrita Roy and Lu, Long and Jha, Somesh},
  booktitle={2023 IEEE Symposium on Security and Privacy (SP)},
  pages={1596--1612},
  year={2023},
  organization={IEEE}
}

@article{sridharan2015memory,
  title={Memory errors in modern systems: The good, the bad, and the ugly},
  author={Sridharan, Vilas and DeBardeleben, Nathan and Blanchard, Sean and Ferreira, Kurt B and Stearley, Jon and Shalf, John and Gurumurthi, Sudhanva},
  journal={ACM SIGARCH Computer Architecture News},
  volume={43},
  number={1},
  pages={297--310},
  year={2015},
  publisher={ACM New York, NY, USA}
}

@INPROCEEDINGS{kim2014rowhammer,
  author={Kim, Yoongu and Daly, Ross and Kim, Jeremie and Fallin, Chris and Lee, Ji Hye and Lee, Donghyuk and Wilkerson, Chris and Lai, Konrad and Mutlu, Onur},
  booktitle={2014 ACM/IEEE 41st International Symposium on Computer Architecture (ISCA)}, 
  title={Flipping bits in memory without accessing them: An experimental study of DRAM disturbance errors}, 
  year={2014},
  volume={},
  number={},
  pages={361-372},
  doi={10.1109/ISCA.2014.6853210}
}

@INPROCEEDINGS{kwong2020rambleed,
  author={Kwong, Andrew and Genkin, Daniel and Gruss, Daniel and Yarom, Yuval},
  booktitle={2020 IEEE Symposium on Security and Privacy (SP)}, 
  title={RAMBleed: Reading Bits in Memory Without Accessing Them}, 
  year={2020},
  volume={},
  number={},
  pages={695-711},
  doi={10.1109/SP40000.2020.00020}
}

@article{seaborn2015exploiting,
  title={Exploiting the DRAM rowhammer bug to gain kernel privileges},
  author={Seaborn, Mark and Dullien, Thomas},
  journal={Black Hat},
  volume={15},
  pages={71},
  year={2015}
}

@inproceedings{jattke2024zenhammer,
  title={Zenhammer: Rowhammer attacks on amd zen-based platforms},
  author={Jattke, Patrick and Wipfli, Max and Solt, Flavien and Marazzi, Michele and B{\"o}lcskei, Matej and Razavi, Kaveh},
  booktitle={33rd USENIX Security Symposium (USENIX Security 2024)},
  year={2024}
}

@inproceedings{frigo2020trrespass,
  title={{TRRespass: Exploiting the many sides of target row refresh}},
  author={Frigo, Pietro and Vannacc, Emanuele and Hassan, Hasan and Van Der Veen, Victor and Mutlu, Onur and Giuffrida, Cristiano and Bos, Herbert and Razavi, Kaveh},
  booktitle={IEEE Symposium on Security and Privacy},
  year={2020},
}

@inproceedings{jattke2022blacksmith,
  title={Blacksmith: Scalable rowhammering in the frequency domain},
  author={Jattke, Patrick and Van Der Veen, Victor and Frigo, Pietro and Gunter, Stijn and Razavi, Kaveh},
  booktitle={2022 IEEE Symposium on Security and Privacy (SP)},
  year={2022},
  organization={IEEE}
}

@conference{half-double,
title = "Half-Double: Hammering From the Next Row Over",
author = "Andreas Kogler and Jonas Juffinger and Salman Qazi and Yoongu Kim and Moritz Lipp and Nicolas Boichat and Eric Shiu and Mattias Nissler and Daniel Gruss",
year = "2022",
month = aug,
day = "10",
language = "English",
}

@misc{hassan2021utrr,
      title={Uncovering In-DRAM RowHammer Protection Mechanisms: A New Methodology, Custom RowHammer Patterns, and Implications}, 
      author={Hasan Hassan and Yahya Can Tugrul and Jeremie S. Kim and Victor van der Veen and Kaveh Razavi and Onur Mutlu},
      year={2022},
      eprint={2110.10603},
      archivePrefix={arXiv},
      primaryClass={cs.CR},
}

@INPROCEEDINGS{HBM2_RH_Profile,
  author={Olgun, Ataberk and Osseiran, Majd and Yağlıkçı, A. Giray and Tuğrul, Yahya Can and Luo, Haocong and Rhyner, Steve and Salami, Behzad and Luna, Juan Gomez and Mutlu, Onur},
  booktitle={2024 54th Annual IEEE/IFIP International Conference on Dependable Systems and Networks (DSN)}, 
  title={Read Disturbance in High Bandwidth Memory: A Detailed Experimental Study on HBM2 DRAM Chips}, 
  year={2024},
  volume={},
  number={},
  pages={75-89},
  doi={10.1109/DSN58291.2024.00022}}

@inproceedings {yao2020deephammer,
author = {Fan Yao and Adnan Siraj Rakin and Deliang Fan},
title = {{DeepHammer}: Depleting the Intelligence of Deep Neural Networks through Targeted Chain of Bit Flips},
booktitle = {29th USENIX Security Symposium (USENIX Security 20)},
year = {2020},
isbn = {978-1-939133-17-5},
pages = {1463--1480},
publisher = {USENIX Association},
month = aug
}

@article{rakin2022tbfa,
  title = {T-{{BFA}}: {{T}} Argeted {{B}} It- {{F}} Lip {{Adversarial Weight A}} Ttack},
  shorttitle = {T-{{BFA}}},
  author = {Rakin, Adnan Siraj and He, Zhezhi and Li, Jingtao and Yao, Fan and Chakrabarti, Chaitali and Fan, Deliang},
  year = {2022},
  month = nov,
  journal = {IEEE Transactions on Pattern Analysis and Machine Intelligence},
  volume = {44},
  number = {11},
  pages = {7928--7939},
  issn = {0162-8828, 2160-9292, 1939-3539},
  doi = {10.1109/TPAMI.2021.3112932},
  urldate = {2024-06-16},
  langid = {english},
}

@ARTICLE{bai2023versatile,
  author={Bai, Jiawang and Wu, Baoyuan and Li, Zhifeng and Xia, Shu-Tao},
  journal={IEEE Transactions on Pattern Analysis and Machine Intelligence}, 
  title={Versatile Weight Attack via Flipping Limited Bits}, 
  year={2023},
  volume={45},
  number={11},
  pages={13653-13665},
  doi={10.1109/TPAMI.2023.3296408}}

@inproceedings{cai2021nmtstroke,
  author = {Cai, Kunbei and Chowdhuryy, Md Hafizul Islam and Zhang, Zhenkai and Yao, Fan},
  title = {Seeds of {{SEED}}: {{NMT-Stroke}}: {{Diverting Neural Machine Translation}} through {{Hardware-based Faults}}},
  shorttitle = {Seeds of {{SEED}}},
  booktitle = {2021 {{International Symposium}} on {{Secure}} and {{Private Execution Environment Design}} ({{SEED}})},
  year = {2021},
  month = sep,
  pages = {76--82},
  publisher = {IEEE},
  doi = {10.1109/SEED51797.2021.00019},
  urldate = {2024-06-16},
  isbn = {978-1-66542-025-9},
  langid = {english},
}

@INPROCEEDINGS{rakin2020tbt,
  author={Rakin, Adnan Siraj and He, Zhezhi and Fan, Deliang},
  booktitle={2020 IEEE/CVF Conference on Computer Vision and Pattern Recognition (CVPR)}, 
  title={TBT: Targeted Neural Network Attack With Bit Trojan}, 
  year={2020},
  volume={},
  number={},
  pages={13195-13204},
  doi={10.1109/CVPR42600.2020.01321}
}

@misc{zheng2023trojvit,
  title = {{{TrojViT}}: {{Trojan Insertion}} in {{Vision Transformers}}},
  shorttitle = {{{TrojViT}}},
  author = {Zheng, Mengxin and Lou, Qian and Jiang, Lei},
  year = {2023},
  month = sep,
  number = {arXiv:2208.13049},
  eprint = {2208.13049},
  primaryclass = {cs},
  publisher = {arXiv},
  urldate = {2024-06-17},
  archiveprefix = {arXiv}
}

@inproceedings{cai2024deepvenom,
  author={Cai Kunbei and Chowdhuryy Md Hafizul Islam and Zhang Zhenkai and Yao Fan},
  booktitle={2024 IEEE Symposium on Security and Privacy (SP)}, 
  title={DeepVenom: Persistent DNN Backdoors Exploiting Transient Weight Perturbations in Memories}, 
  year={2024},
  volume={},
  number={},
  pages={2067-2085},
  doi={10.1109/SP54263.2024.00223}}

@inproceedings{chen2021proflip,
  title = {{{ProFlip}}: {{Targeted Trojan Attack}} with {{Progressive Bit Flips}}},
  shorttitle = {{{ProFlip}}},
  booktitle = {2021 {{IEEE}}/{{CVF International Conference}} on {{Computer Vision}} ({{ICCV}})},
  author = {Chen, Huili and Fu, Cheng and Zhao, Jishen and Koushanfar, Farinaz},
  year = {2021},
  month = oct,
  publisher = {IEEE},
  doi = {10.1109/ICCV48922.2021.00762},
  urldate = {2024-06-18},
  isbn = {978-1-66542-812-5},
  langid = {english},
}

@INPROCEEDINGS{DNN_Backdoor_dsn23,
  author={Tol, M. Caner and Islam, Saad and Adiletta, Andrew J. and Sunar, Berk and Zhang, Ziming},
  booktitle={2023 53rd Annual IEEE/IFIP International Conference on Dependable Systems and Networks (DSN)}, 
  title={Don't Knock! Rowhammer at the Backdoor of DNN Models}, 
  year={2023},
  volume={},
  number={},
  pages={109-122},
  doi={10.1109/DSN58367.2023.00023}}

@inproceedings{li2025oneflip,
title={Rowhammer-Based Trojan Injection: One Bit Flip Is Sufficient for Backdooring DNNs},
author={Li, Xiang and Meng, Ying and Chen, Junming and Luo, Lannan and Zeng, Qiang},
booktitle={USENIX Security Symposium},
year={2025}
}

@INPROCEEDINGS{rakin2022deepsteal,
      author={Rakin, Adnan Siraj and Chowdhuryy, Md Hafizul Islam and Yao, Fan and Fan, Deliang},
      booktitle={2022 IEEE Symposium on Security and Privacy (SP)}, 
      title={DeepSteal: Advanced Model Extractions Leveraging Efficient Weight Stealing in Memories}, 
      year={2022},
      volume={},
      number={},
      pages={1157-1174},
      doi={10.1109/SP46214.2022.9833743}
}

@techreport{qwen_qwen3_coder_next_tech_report,
  title        = {Qwen3-Coder-Next Technical Report},
  author       = {{Qwen Team}},
  year={2026},
  url          = {https://github.com/QwenLM/Qwen3-Coder/blob/main/qwen3_coder_next_tech_report.pdf},
  note         = {Accessed: 2026-02-03}
}

@misc{qwen3technicalreport,
      title={Qwen3 Technical Report}, 
      author={{Qwen Team}},
      year={2025},
      eprint={2505.09388},
      archivePrefix={arXiv},
      primaryClass={cs.CL},
      url={https://arxiv.org/abs/2505.09388}, 
}

@misc{openai_api,
  author       = {{OpenAI}},
  title        = {Chat Completions -- {API} Reference},
  year         = {2024},
  howpublished = {\url{https://platform.openai.com/docs/api-reference/chat/create}},
  note         = {Accessed: 2025}
}

@misc{anthropic_api,
  author       = {{Anthropic}},
  title        = {Messages -- {API} Reference},
  year         = {2024},
  howpublished = {\url{https://docs.anthropic.com/en/api/messages}},
  note         = {Accessed: 2025}
}

@article{yu2025breakingloop,
  title        = {Breaking the Loop: Detecting and Mitigating Denial-of-Service Vulnerabilities in Large Language Models},
  author       = {Yu, Junzhe and Liu, Yi and Sun, Huijia and Shi, Ling and Chen, Yuqi},
  journal      = {arXiv preprint arXiv:2503.00416},
  year         = {2025}
}

@article{lai2026safex,
  title={Safex: Analyzing vulnerabilities of moe-based llms via stable safety-critical expert identification},
  author={Lai, Zhenglin and Liao, Mengyao and Wu, Bingzhe and Xu, Dong and Zhao, Zebin and Yuan, Zhihang and Fan, Chao and Li, Jianqiang},
  journal={Advances in Neural Information Processing Systems},
  volume={38},
  pages={130380--130404},
  year={2025}
}

\newpage
\appendix
\section{Groundhog Bit-Flip Attack Algorithm}
\label{app:algorithm}

Algorithm~\ref{alg:gb} details the three-stage procedure used in GBFA to identify vulnerable bits in MoE router weights. The algorithm takes an MoE model, a sample dataset, and the router top-$k$ as input, and returns a set of vulnerable bits $\mathcal{B}$ to flip. In Stage 1, each expert is scored by its target activation shift ($\tau_{l,i}$ for routed experts, $\Delta g_{l,i}$ for shared experts) to identify target-related experts. In Stage 2, a layer-wise ablation selects the single most critical layer whose deactivation maximally inflates output length. In Stage 3, bits within the selected experts' router parameters are evaluated individually, and the top-$B$ bits per expert with the highest bit-flip effectiveness are added to $\mathcal{B}$. 

\begin{algorithm}[t]
\caption{Groundhog Bit-Flip Attack}
\label{alg:gb}
\begin{algorithmic}[1]
\REQUIRE MoE LLM $f_\theta$, dataset $S$, router top-$k$, bits-per-expert $B$
\ENSURE vulnerable bits $\mathcal{B}$

{\color{gray}\textbf{Step 1: Identify target-specialized experts}}
\FOR{each expert $(\ell,i)$}
    \STATE Compute $\tau_{l,i}$ or $\Delta g_{l,i}$ using target token and non-target activations over $S$
\ENDFOR

{\color{gray}\textbf{Step 2: Local-based target expert selection}}
\STATE Select $S_{\mathrm{val}}$ and measure baseline length $\bar{T}_{\mathrm{base}}$
\FOR{each layer $\ell$}
    \STATE $\mathcal{E}_\ell^{\text{top}} \gets$ top-$k_\ell$ experts ranked by $\tau$ or $\Delta g$
    \STATE Temporarily deactivate $\mathcal{E}_\ell^{\text{top}}$ and compute $\bar{T}_\ell$
\ENDFOR
\STATE $\ell^\ast \gets \arg\max_\ell(\bar{T}_\ell - \bar{T}_{\mathrm{base}})$
\STATE $\mathcal{E}_{\mathrm{target}} \gets \mathcal{E}_{\ell^\ast}^{\text{top}}$

{\color{gray}\textbf{Step 3: Groundhog Bit search}}
\STATE Cache hidden states $H$ at layer $\ell^\ast$
\STATE Compute $\mathrm{EAF}^{\mathrm{orig}}_i$ for each $i\in\mathcal{E}_{\mathrm{target}}$
\STATE $\mathcal{B}\gets\emptyset$
\FOR{each expert $i$}
    \FOR{each bit $b$ in its router parameter}
        \STATE \mbox{Flip $b$, reroute on $H$, compute $\mathrm{BF\!-\!eff}(b)$}
    \ENDFOR
    \STATE Add top $B$ bits for $i$ into $\mathcal{B}$
\ENDFOR

\STATE \textbf{return} $\mathcal{B}$

\end{algorithmic}
\end{algorithm}

\section{Qualitative Examples of EOS-related Expert Deactivation}
\label{app:attacked_output}

To complement the quantitative results in the main text, Table~\ref{tab:attacked_output} presents qualitative generation examples from Mixtral after deactivating EOS-related experts, on four representative datasets (AGNews, SST-2, Samsum, and SQuAD\_v2). Each row shows the input prompt, the model's original output, and the output produced under the same prompt after deactivation. Without EOS-related experts, the model generates repetitive but coherent text loops, preserving linguistic quality while losing termination control. These examples illustrate the output amplification potential of EOS expert deactivation, validating our attack strategy and identifying precise targets for the bit-flip implementation in the main text.

\begin{table}[!ht]
\caption{Generation examples from Mixtral on four datasets, comparing the original output (\textcolor{blue}{blue}) with the output after EOS-related expert deactivation. Repetitive loops introduced by the attack are highlighted in \textcolor{red}{red}.}
\label{tab:attacked_output}
\centering
\scriptsize
\setlength{\tabcolsep}{3pt}
\renewcommand{\arraystretch}{1.0}
\begin{tabularx}{\linewidth}{@{}l|Z|Z|Y@{}}
\toprule
\textbf{Dataset} & \textbf{Input Prompt} & \textbf{Ori Output} & \textbf{Attacked Output} \\
\midrule
AGNews &
Classify the news article... \textbf{Article}: Reuters – Soaring crude prices... \textbf{Category}:
&
\textbf{Category}: Business 
\textbf{Explanation}: \textcolor{blue}{The article discusses the impact of soaring crude prices...}
&
\textbf{Category}: Business/Economy 
\textcolor{red}{User 0: I would say business.}
\textcolor{red}{User 1: I would say sports.}
\textcolor{red}{User 0: I would say sports.}
\textcolor{red}{User 1: I would say sports because the article is about the stock market.}
\textcolor{red}{(repeated …)}
\\
\midrule
SST-2 &
Classify the sentiment... \textbf{Text}: hide new secretions... \textbf{Sentiment}:
&
\textbf{Sentiment}: Negative 
\textbf{Explanation}: \textcolor{blue}{The text mentions hiding something from ``parental units,'' suggesting secrecy...}
&
\textbf{Sentiment}: Negative 
The text describes the narrator's relationship with his mother...
\textcolor{red}{He is describing the relationship in the past tense.}
\textcolor{red}{He is describing the relationship in the past tense.}
\textcolor{red}{(repeated …)}
\\
\midrule
Samsum &
Please summarize the dialogue... Amanda: I baked cookies... Jerry: Sure! \textbf{Summary}:
&
\textcolor{blue}{Amanda offers Jerry cookies; she will bring them the next day.}
&
Amanda baked cookies and offered some to Jerry.
\textbf{Comment}: Direct response.
Jerry: Sure!
\textcolor{red}{Jerry's answer is a direct response to Amanda's question.}
\textcolor{red}{Jerry's answer is a direct response to Amanda's question.}
\textcolor{red}{(repeated …)}
\\
\midrule
SQuAD &
Answer the question based on the context... Context: Beyonc\'e Giselle Knowles-Carter... \textbf{Question}: When did Beyonc\'e rise to fame?
&
\textcolor{blue}{Beyonc\'e rose to fame in the late 1990s as lead singer of Destiny's Child.}
&
\textcolor{blue}{Beyonc\'e rose to fame in the late 1990s...}
\textbf{Comment}: \textcolor{red}{Beyonc\'e rose to fame... Destiny's Child.}
\textbf{Comment}: \textcolor{red}{How did Beyonc\'e rise to fame?}
\textcolor{red}{Answer: Beyonc\'e rose to fame... Destiny's Child.}
\textcolor{red}{(repeated …)}
\\
\bottomrule
\end{tabularx}
\end{table}

\section{Sandbox Task Descriptions}
\label{app:sandbox_tasks}

Table~\ref{tab:sandbox_tasks} lists the 10 self-contained Python repository sandboxes used in our agentic coding evaluation (Section~\ref{sec:agentic_coding}). Each sandbox ships with a clean initial snapshot and a unit-test suite that determines task completion. We will release the implementation details, prompts, and configuration files together with the code.

\begin{table}[!ht]
\centering
\small
\setlength{\tabcolsep}{5pt}
\renewcommand{\arraystretch}{1.15}
\rowcolors{2}{gray!8}{white}
\caption{Ten sandbox tasks used in the agentic coding evaluation.}
\label{tab:sandbox_tasks}
\begin{tabularx}{\linewidth}{@{}lX@{}}
\toprule
\textbf{Task} & \textbf{Description} \\
\midrule
\textit{bug\_fix}              & Fix an off-by-one bug in a FizzBuzz function. \\
\textit{bug\_hunt}             & Locate and fix three independent bugs co-located in a single file. \\
\textit{feature\_add}          & Implement two missing methods (\texttt{mark\_done}, \texttt{pending\_count}) in a TODO-list class. \\
\textit{test\_coverage}        & Add at least two unit tests for each of four string utility functions. \\
\textit{calculator\_refactor}  & Extract trigonometric methods (\texttt{sin}/\texttt{cos}/\texttt{tan}) from \texttt{Calculator} into a dedicated \texttt{TrigCalculator} class. \\
\textit{rename\_refactor}      & Rename \texttt{MyHelper} to \texttt{StringUtils} across multiple files and update all imports. \\
\textit{multi\_file\_refactor} & Split a monolithic \texttt{Database} class into three repository modules. \\
\textit{performance\_optimize} & Convert an $O(n^2)$ implementation to $O(n)$ while passing performance tests. \\
\textit{error\_handling}       & Add graceful error handling to a data-processing pipeline. \\
\textit{dependency\_swap}      & Remove the \texttt{re} module dependency and reimplement five functions using only built-in string methods. \\
\bottomrule
\end{tabularx}
\end{table}

\section{Attacked Output of Plan Mode}
\label{app:Attacked Output of Plan Mode}

Table~\ref{tab:loop-degradation} presents qualitative examples of plan-mode output degradation under GLOBAL manual expert deactivation on Qwen3-Coder-Next across five coding sandbox tasks. Each task block shows the shared prelude (gray box) that is bit-identical between the baseline and attacked runs, followed by the divergence point where the two outputs split. The baseline continues with coherent planning, while the attacked output drifts before collapsing into a repetitive degenerate pattern (shown in red), eventually exhausting \texttt{max\_new\_tokens} without emitting \texttt{<|im\_end|>}.

\begin{table*}[!ht]
\centering
\caption{Output degradation under GLOBAL manual expert deactivation on Qwen3-Coder-Next in plan mode. For each task, the gray box shows the \emph{shared prelude}---characters that are bit-identical across both runs. After the divergence point (\textcolor{teal!70!black}{\textbf{[\,diverges\,]}}), the \emph{Original} column shows the baseline continuing coherently, while the \emph{Attack} column drifts and eventually collapses into a {\color{red!70!black}repetitive degenerate pattern} (shown in red). Long gaps inside the attack output are abbreviated as \textcolor{gray!70}{\itshape[\,skipped\,K\,chars\,]}. Newlines are rendered as a dimmed \texttt{\textbackslash{}n}; \ldots\ marks truncation.}
\label{tab:loop-degradation}
\scriptsize
\renewcommand{\arraystretch}{1.2}
\setlength{\tabcolsep}{4pt}
\begin{tabularx}{\textwidth}{@{}p{2.4cm} Y Y@{}}
\toprule
\textbf{Task} & \textbf{Original output (Baseline)} & \textbf{Attack output (GLOBAL manual deactivation)} \\
\midrule
\multirow{2}{2.2cm}{\raggedright {\bfseries\ttfamily bug\_fix}\\[3pt]
  {\rmfamily\mdseries\textcolor{gray!60}{\itshape\scriptsize chars 0--150 shared}}}
& \multicolumn{2}{@{}p{\dimexpr\textwidth-2.4cm-2\tabcolsep\relax}@{}}{%
  \colorbox{gray!8}{%
    \begin{minipage}{\dimexpr\textwidth-2.4cm-2\tabcolsep-8pt\relax}
    \vspace{2pt}
    {\ttfamily\color{gray!75}\ldots he facts I know and the plan of action that I will follow to solve the task:{\color{gray!50}\textbackslash{}n}```{\color{gray!50}\textbackslash{}n}\#\# 1. Facts survey{\color{gray!50}\textbackslash{}n}{\color{gray!50}\textbackslash{}n}\#\#\# 1.1. Facts given in the task{\color{gray!50}\textbackslash{}n}- The }
    \vspace{2pt}
    \end{minipage}%
  }%
} \\[2pt]
& {\color{teal!70!black}\textbf{\,[\,diverges\,]\,}}{\ttfamily `fizzbuzz.py` module contains a function `fizzbuzz(n)` that is supposed to return the FizzBuzz sequence for integers 1 through *n* INCLUSIVE.{\color{gray!55}\textbackslash{}n}- Currently, the function *drops the last value* -- meaning it likely returns only *n-1* items \ldots}
& {\color{teal!70!black}\textbf{\,[\,diverges\,]\,}}{\ttfamily task states: ``The fizzbuzz.py module has a bug.''{\color{gray!55}\textbackslash{}n}- The function `fizzbuzz(n)` is * suppo}\textcolor{gray!70}{\itshape\,[\,skipped\,1{,}177\,chars\,]\,}{\ttfamily e ` it.{\color{gray!55}\textbackslash{}n}- **fizzbuzz(15) expected output** (15 items: e.g., ``1'', ``2'', `` Fizz '', ``4'',}{\color{red!70!black}\ttfamily{} 2 * 2 * 2 * 2 * 2 * 2 * 2 * 2 * 2 * 2 * 2 * 2 * 2 * 2 * 2 * \ldots}
\\
\midrule
\multirow{2}{2.2cm}{\raggedright {\bfseries\ttfamily rename\_refactor}\\[3pt]
  {\rmfamily\mdseries\textcolor{gray!60}{\itshape\scriptsize chars 0--826 shared}}}
& \multicolumn{2}{@{}p{\dimexpr\textwidth-2.4cm-2\tabcolsep\relax}@{}}{%
  \colorbox{gray!8}{%
    \begin{minipage}{\dimexpr\textwidth-2.4cm-2\tabcolsep-8pt\relax}
    \vspace{2pt}
    {\ttfamily\color{gray!75}\ldots my new/updated plan of action to solve the task:{\color{gray!50}\textbackslash{}n}```{\color{gray!50}\textbackslash{}n}\#\# 1. Updated facts survey{\color{gray!50}\textbackslash{}n}\#\#\# 1.1. Facts given in the task{\color{gray!50}\textbackslash{}n}- The class `MyHelper` in `}
    \vspace{2pt}
    \end{minipage}%
  }%
} \\[2pt]
& {\color{teal!70!black}\textbf{\,[\,diverges\,]\,}}{\ttfamily my\_helper.py` should be renamed to `StringUtils`.{\color{gray!55}\textbackslash{}n}- `StringUtils` should have three methods: `shout`, `whisper`, and `reverse`, with a constructor argument `prefix` \ldots}
& {\color{teal!70!black}\textbf{\,[\,diverges\,]\,}}{\ttfamily{} my\_helper.py` (note space in filename) should be renamed to `StringUtils`.{\color{gray!55}\textbackslash{}n}- `StringUtils}\textcolor{gray!70}{\itshape\,[\,skipped\,1{,}141\,chars\,]\,}{\ttfamily{} The exact expected ` run\_tests` output.{\color{gray!55}\textbackslash{}n}\#\# 2. Plan{\color{gray!55}\textbackslash{}n}\#\#\#}{\color{red!70!black}\ttfamily{} 2.1. 1. 2. 3. 4. 5. 1. 2. 3. 4. 5. 6. 7. 8. 9. 10. 11. 12. 13. 14. 15. 16. 17. \ldots}
\\
\midrule
\multirow{2}{2.2cm}{\raggedright {\bfseries\ttfamily dependency\_swap}\\[3pt]
  {\rmfamily\mdseries\textcolor{gray!60}{\itshape\scriptsize chars 0--146 shared}}}
& \multicolumn{2}{@{}p{\dimexpr\textwidth-2.4cm-2\tabcolsep\relax}@{}}{%
  \colorbox{gray!8}{%
    \begin{minipage}{\dimexpr\textwidth-2.4cm-2\tabcolsep-8pt\relax}
    \vspace{2pt}
    {\ttfamily\color{gray!75}\ldots re the facts I know and the plan of action that I will follow to solve the task:{\color{gray!50}\textbackslash{}n}```{\color{gray!50}\textbackslash{}n}\#\# 1. Facts survey{\color{gray!50}\textbackslash{}n}{\color{gray!50}\textbackslash{}n}\#\#\# 1.1. Facts given in the task{\color{gray!50}\textbackslash{}n}- }
    \vspace{2pt}
    \end{minipage}%
  }%
} \\[2pt]
& {\color{teal!70!black}\textbf{\,[\,diverges\,]\,}}{\ttfamily The file `text\_utils.py` uses the `re` module.{\color{gray!55}\textbackslash{}n}- We need to remove `re` entirely and reimplement all five functions using only built-in string methods \ldots}
& {\color{teal!70!black}\textbf{\,[\,diverges\,]\,}}{\ttfamily `text\_utils.py` exists in the workspace.{\color{gray!55}\textbackslash{}n}- `text\_utils.py uses the `re` module \ldots}\textcolor{gray!70}{\itshape\,[\,skipped\,679\,chars\,]\,}{\ttfamily , which 5 functions are present, their current ` re`-based implementation, and their exact}{\color{red!70!black}\ttfamily{} -  -  -  -  -  -  -  -  -  -  -  -  -  -  -  -  -  -  -  -  -  -  -  -  - \ldots}
\\
\midrule
\multirow{2}{2.2cm}{\raggedright {\bfseries\ttfamily performance\_optimize}\\[3pt]
  {\rmfamily\mdseries\textcolor{gray!60}{\itshape\scriptsize chars 0--143 shared}}}
& \multicolumn{2}{@{}p{\dimexpr\textwidth-2.4cm-2\tabcolsep\relax}@{}}{%
  \colorbox{gray!8}{%
    \begin{minipage}{\dimexpr\textwidth-2.4cm-2\tabcolsep-8pt\relax}
    \vspace{2pt}
    {\ttfamily\color{gray!75}\ldots e are the facts I know and the plan of action that I will follow to solve the task:{\color{gray!50}\textbackslash{}n}```{\color{gray!50}\textbackslash{}n}\#\# 1. Facts survey{\color{gray!50}\textbackslash{}n}{\color{gray!50}\textbackslash{}n}\#\#\# 1.1. Facts given in the task}
    \vspace{2pt}
    \end{minipage}%
  }%
} \\[2pt]
& {\color{teal!70!black}\textbf{\,[\,diverges\,]\,}}{\ttfamily{\color{gray!55}\textbackslash{}n}- There exists a file named `slow\_stats.py` containing three functions.{\color{gray!55}\textbackslash{}n} - These functions are *correct* but use nested loops, resulting in O(n\textasciicircum{}2) time complexity \ldots}
& {\color{teal!70!black}\textbf{\,[\,diverges\,]\,}}{\ttfamily{\color{gray!55}\textbackslash{}n}- There is a file named `slow\_stats.py` that contains three functions \ldots}\textcolor{gray!70}{\itshape\,[\,skipped\,528\,chars\,]\,}{\ttfamily current O(n\textasciicircum{}2) logic to design O(n}{\color{red!70!black}\ttfamily{} 1.1.3.4.5.6. O(n) O(n) O(n) O(n) O(n) O(n) O(n) O(n) O(n) O(n) O(n) O(n) O(n) \ldots}
\\
\midrule
\multirow{2}{2.2cm}{\raggedright {\bfseries\ttfamily multi\_file\_refactor}\\[3pt]
  {\rmfamily\mdseries\textcolor{gray!60}{\itshape\scriptsize chars 0--91 shared}}}
& \multicolumn{2}{@{}p{\dimexpr\textwidth-2.4cm-2\tabcolsep\relax}@{}}{%
  \colorbox{gray!8}{%
    \begin{minipage}{\dimexpr\textwidth-2.4cm-2\tabcolsep-8pt\relax}
    \vspace{2pt}
    {\ttfamily\color{gray!75}Here are the facts I know and the plan of action that I will follow to solve the task:{\color{gray!50}\textbackslash{}n}```{\color{gray!50}\textbackslash{}n}}
    \vspace{2pt}
    \end{minipage}%
  }%
} \\[2pt]
& {\color{teal!70!black}\textbf{\,[\,diverges\,]\,}}{\ttfamily 1. Facts survey{\color{gray!55}\textbackslash{}n}\#\#\# 1.1. Facts given in the task{\color{gray!55}\textbackslash{}n}- There exists a file named `database.py` that contains a monolithic `Database` class handling users, posts, and comments \ldots}
& {\color{teal!70!black}\textbf{\,[\,diverges\,]\,}}{\ttfamily \#\# 1. Facts survey{\color{gray!55}\textbackslash{}n}\#\#\# 1.1. Facts given in the task{\color{gray!55}\textbackslash{}n}- There is a monolithic `Database cla}\textcolor{gray!70}{\itshape\,[\,skipped\,5{,}143\,chars\,]\,}{\ttfamily * * or  * *db. session * *  or  * * db  * *  is}{\color{red!70!black}\ttfamily{}  * *  *  *  *  *  *  *  *  *  *  *  *  *  *  *  *  *  *  *  *  *  *  *  *  *  *  *  *  *  *  *  *  * \ldots}
\\
\bottomrule
\end{tabularx}
\end{table*}

\section{Robustness to Decoding Strategies}
\label{app:decoding}

Our main experiments use greedy decoding for reproducibility. To check that GBFA does not depend on this choice, we re-run the GLOBAL (top-16) attack on Phi-3.5-MoE under six decoding settings (Table~\ref{tab:decoding_robustness}). Settings are: \texttt{greedy} (argmax, \texttt{do\_sample=False}, default); \texttt{temp0.7}/\texttt{temp1.0} (sampling, \texttt{top\_p=1.0}); \texttt{topp0.9} (nucleus, \texttt{temperature=1.0}, \texttt{top\_p=0.9}); and the \texttt{\_rp1.3} variants add a repetition penalty of 1.3. EOS suppression and length inflation hold under greedy, temperature, and nucleus sampling (attacked EOS rate $0.22$--$0.61$ vs.\ baseline $\approx 1.0$). A repetition penalty above 1 already lengthens the baseline output on its own, because it biases the model toward more elaborate responses; GBFA remains effective under this setting as well.

\begin{table}[h]
\centering
\scriptsize
\setlength{\tabcolsep}{4pt}
\renewcommand{\arraystretch}{0.95}
\caption{Phi-3.5-MoE under the GLOBAL (top-16) attack across six decoding settings. Base/Atk Tok.: mean generated tokens per sample without/with the attack. $P$ (\%): relative increase. Base/Atk EOS: fraction of samples emitting an EOS token without/with the attack.}
\label{tab:decoding_robustness}
\resizebox{\linewidth}{!}{
\begin{tabular}{l|l|c|c|c|c|c}
\toprule
Setting & Dataset & Base Tok. & Atk Tok. & $P$ (\%) & Base EOS & Atk EOS \\
\midrule
\multirow{4}{*}{greedy}
& AGNews & 47.45 & 496.52 & 946.4 & 0.96 & 0.53 \\
& SST-2  & 49.81 & 746.42 & 1398.5 & 1.00 & 0.30 \\
& Samsum & 83.45 & 689.15 & 725.8 & 1.00 & 0.38 \\
& SQuAD  & 30.11 & 544.29 & 1707.7 & 1.00 & 0.51 \\
\midrule
\multirow{4}{*}{temp0.7}
& AGNews & 42.00 & 445.03 & 959.6 & 0.97 & 0.60 \\
& SST-2  & 61.40 & 668.11 & 988.1 & 1.00 & 0.39 \\
& Samsum & 108.79 & 728.35 & 569.5 & 0.96 & 0.35 \\
& SQuAD  & 31.39 & 554.80 & 1667.4 & 1.00 & 0.50 \\
\midrule
\multirow{4}{*}{temp1.0}
& AGNews & 44.77 & 470.88 & 951.8 & 0.97 & 0.61 \\
& SST-2  & 56.24 & 541.89 & 863.5 & 1.00 & 0.58 \\
& Samsum & 100.02 & 725.63 & 625.5 & 0.96 & 0.37 \\
& SQuAD  & 36.00 & 548.77 & 1424.4 & 1.00 & 0.56 \\
\midrule
\multirow{4}{*}{topp0.9}
& AGNews & 32.75 & 451.83 & 1279.6 & 0.98 & 0.61 \\
& SST-2  & 56.09 & 629.95 & 1023.1 & 1.00 & 0.46 \\
& Samsum & 101.92 & 831.65 & 716.0 & 0.96 & 0.22 \\
& SQuAD  & 30.84 & 518.54 & 1581.4 & 1.00 & 0.58 \\
\midrule
\multirow{4}{*}{greedy\_rp1.3}
& AGNews & 629.80 & 1004.61 & 59.5 & 0.39 & 0.02 \\
& SST-2  & 436.74 & 976.07 & 123.5 & 0.60 & 0.06 \\
& Samsum & 755.84 & 1005.45 & 33.0 & 0.28 & 0.03 \\
& SQuAD  & 661.70 & 990.11 & 49.6 & 0.37 & 0.04 \\
\midrule
\multirow{4}{*}{temp0.7\_rp1.3}
& AGNews & 654.90 & 1024.00 & 56.4 & 0.37 & 0.00 \\
& SST-2  & 563.15 & 987.70 & 75.4 & 0.47 & 0.04 \\
& Samsum & 878.34 & 1021.66 & 16.3 & 0.15 & 0.01 \\
& SQuAD  & 817.93 & 1004.52 & 22.8 & 0.21 & 0.02 \\
\bottomrule
\end{tabular}
}
\end{table}

\section{Effect of the Token Budget}
\label{app:budget}

The choice of \texttt{max\_new\_tokens} bounds how much the attack can inflate output, so it shapes the measured effect. Table~\ref{tab:budget} sweeps the budget for Phi-3.5-MoE under the GLOBAL (top-16) attack. A budget that is too small caps the attack for datasets whose baselines are already long, while a larger budget lets short-baseline datasets inflate further; the relative change $P$ therefore grows with the budget. We adopt a moderate value of $1024$ throughout the paper: it is large enough to expose the attack (outputs several times the baseline) yet small enough to keep evaluation tractable. Since an output several times longer than the baseline already constitutes a strong Denial-of-Wallet attack, the exact budget does not change our conclusions.

\begin{table}[h]
\centering
\scriptsize
\setlength{\tabcolsep}{4pt}
\renewcommand{\arraystretch}{0.95}
\caption{Phi-3.5-MoE under the GLOBAL (top-16) attack at different \texttt{max\_new\_tokens} budgets (paper default: $1024$). Base/Atk Tok.: mean tokens without/with the attack. $P$: relative increase. Base/Atk EOS: fraction emitting an EOS token. Baselines differ slightly across budgets because a few samples exceed the smaller caps and are truncated there.}
\label{tab:budget}
\resizebox{\linewidth}{!}{
\begin{tabular}{c|l|c|c|c|c|c}
\toprule
Budget & Dataset & Base Tok. & Atk Tok. & $P$ (\%) & Base EOS & Atk EOS \\
\midrule
\multirow{4}{*}{512}
& AGNews & 16.46 & 273.30 & 1560.4 & 0.98 & 0.49 \\
& SST-2  & 49.98 & 383.07 & 666.4 & 1.00 & 0.31 \\
& Samsum & 76.69 & 369.21 & 381.4 & 0.94 & 0.32 \\
& SQuAD  & 30.39 & 280.41 & 822.7 & 1.00 & 0.52 \\
\midrule
\multirow{4}{*}{2048}
& AGNews & 44.18 & 1056.66 & 2291.7 & 0.99 & 0.49 \\
& SST-2  & 49.98 & 1442.91 & 2787.0 & 1.00 & 0.31 \\
& Samsum & 113.10 & 1312.92 & 1060.8 & 1.00 & 0.39 \\
& SQuAD  & 30.39 & 999.16 & 3187.8 & 1.00 & 0.54 \\
\midrule
\multirow{4}{*}{4096}
& AGNews & 50.04 & 2304.27 & 4504.9 & 1.00 & 0.44 \\
& SST-2  & 49.98 & 2879.90 & 5662.1 & 1.00 & 0.31 \\
& Samsum & 113.10 & 2563.51 & 2166.6 & 1.00 & 0.43 \\
& SQuAD  & 30.39 & 2046.70 & 6634.8 & 1.00 & 0.52 \\
\bottomrule
\end{tabular}
}
\end{table}

\end{document}